\documentclass[twocolumn, switch]{article}
\usepackage{arxiv}

\usepackage[utf8]{inputenc}

\usepackage{hyperref}
\usepackage{xstring}
\usepackage{color}
\usepackage[numbers,square]{natbib}
\usepackage{url}
\usepackage{graphicx}
\usepackage{amsmath,amssymb,amsfonts,array}
\usepackage[justification=centering]{caption} % hide
\usepackage{subcaption}
\usepackage{multirow}
\usepackage{booktabs,threeparttable}

\usepackage{tabularx}

\usepackage{float}
\usepackage{lipsum}
\usepackage{authblk}

\usepackage{utf8}
\novocalize
\setfarsi
\settrans{farsi}
\yahnodots

\newsavebox{\spacebox}
\begin{lrbox}{\spacebox}
\verb*! !
\end{lrbox}
\newcolumntype{L}{>{\raggedright\arraybackslash}X}
\newcommand{\filliation}[5]{\affil[#1]{\textbf{#2}\vskip 0pt \textbf{#3}\vskip 0pt #4\vskip 0pt #5\vspace{10pt}}} %% 5 number of affil 1

\title{ParsBERT: Transformer-based Model for Persian Language Understanding}

\author[1]{Mehrdad Farahani}
\author[2]{Mohammad Gharachorloo}
\author[3]{Marzieh Farahani}
\author[4]{Mohammad Manthouri}

\filliation{1}{Department of Computer Engineering}{Islamic Azad University North Tehran Branch}{Tehran, Iran}{m.farahani@iau-tnb.ac.ir}
\filliation{2}{School of Electrical Engineering and Robotics}{Queensland University of Technology}{Brisbane, Australia}{mohammad.gharachorloo@connect.qut.edu.au}
\filliation{3}{Department of Computing Science}{Umeå University}{Umeå, Sweden}{mafa2431@student.umu.se}
\filliation{4}{Department of Electrical and Electronic Engineering}{Shahed Univerisity}{Tehran, Iran}{mmanthouri@shahed.ac.ir}

\begin{document}

\twocolumn[ % Method A for two-column formatting
  \begin{@twocolumnfalse} % Method A for two-column formatting
  
\maketitle

\begin{abstract}
The surge of pre-trained language models has begun a new era in the field of Natural Language Processing (NLP) by allowing us to build powerful language models. Among these models, Transformer-based models such as BERT have become increasingly popular due to their state-of-the-art performance. However, these models are usually focused on English, leaving other languages to multilingual models with limited resources. This paper proposes a monolingual BERT for the Persian language (ParsBERT), which shows its state-of-the-art performance compared to other architectures and multilingual models. Also, since the amount of data available for NLP tasks in Persian is very restricted, a massive dataset for different NLP tasks as well as pre-training the model is composed. ParsBERT obtains higher scores in all datasets, including existing ones as well as composed ones and improves the state-of-the-art performance by outperforming both multilingual BERT and other prior works in Sentiment Analysis, Text Classification and Named Entity Recognition tasks. 
\end{abstract}
\vspace{0.35cm}
\end{@twocolumnfalse}
]

\section{Introduction}
Natural language is the tool humans use to communicate with each other. Thus, a vast amount of data is encoded as texts using this tool. Extracting meaningful information from this type of data and manipulating them using computers lie within the field of Natural Language Processing (NLP). There are different NLP tasks such as Named Entity Recognition (NER), Sentiment Analysis (SA), and Question/Answering, each focusing on a particular aspect of the text data to achieve successful performance on each of these tasks, a variety of pre-trained word embedding and language modeling methods have been proposed in the recent years.

Word2Vec \cite{mikolov2013distributed} and GloVe \cite{pennington2014glove} are pre-trained word embeddings methods based on Neural Networks (NNs) that investigate the semantic, syntactic, and logical relationships between words in a sequence to provide a static word representation vectors, based on the training data. While these methods leave the context of the input sequence out of the equation, contextualized word embedding methods such as ELMo \cite{peters2018deep} provide dynamic word embeddings by taking the context into account.

There are two approaches towards pre-trained language representations \cite{devlin2018bert}: feature-based such as ELMo and fine-tuning such as OpenAI GPT \cite{radford2018improving}. Fine-tuning approaches (also known as Transfer Learning methods) seek to train a language model with large datasets of unlabeled plain texts. The parameters of these models are then fine-tuned using task-specific data to achieve state-of-the-art performance over various NLP tasks \cite{devlin2018bert,radford2018improving,liu2019roberta}. The fine-tuning phase, relative to pre-training, requires much less energy and time. Therefore, pre-trained language models can be used to save energy, time, and cost. However, this comes with specific challenges. The amount of data and the computational resources required to pre-train an efficient language model with acceptable performance is substantial; hundreds of gigabytes of text-documents and hundreds of Graphical Processing Units (GPUs) \cite{yang2019xlnet,liu2019roberta,raffel2019exploring,conneau2019unsupervised}.

As a solution, multilingual models have been developed, which can be beneficial for languages with similar morphology and syntactic structure (e.g., Latin-based languages). Other Non-Latin languages differ from Latin-based languages significantly and can not benefit from their shared representations. Therefore, a language-specific approach should be adapted. For instance, the framework of Recurrent Neural Network (RNN), along with morpheme representation, is proposed to overcome feature engineering and data sparsity for the Mongolian NER task \cite{wang2019learning}.

A similar situation applies to the Persian language. Although some multilingual models include Persian, they are susceptible to fall behind monolingual models that are concretely trained over language-specific vocabulary with more massive amounts of Persian text data. To the best of our knowledge, no specific effort has been made to pre-train a Bidirectional Encoder Representation Transformer (BERT) \cite{devlin2018bert} model for the Persian language.

In this paper, we take advantage of the BERT architecture \cite{devlin2018bert} to build a pre-trained language model for the Persian Language, which we call ParsBERT hereafter. We evaluate this model on three Persian NLP downstream tasks: (a) Sentiment Analysis, (b) Text Classification, and (c) Named Entity Recognition. We show that for all these tasks, ParsBERT outperforms several baselines, including previous multilingual and monolingual models. Thus, our contribution can be summarized as follows:

\begin{itemize}
\item Proposing a monolingual Persian language model (ParsBERT) based on the BERT architecture.
\item ParsBERT achieves better performances regarding other multilingual and deep-hybrid architectures.
\item ParsBERT is lighter than the original multilingual BERT model.
\item During this procedure, the research provided a massive set of Persian text corpora and NLP tasks for other uses cases.
\end{itemize}

The rest of this paper is organized as follows. In section \ref{sec:related}, a comprehensive study of previous related works is provided. Section \ref{sec:methodology} outlines the methodology used to pre-train ParsBERT. In the next section, \ref{sec:evaluation} describes the NLP downstream tasks and benchmark datasets on which the model is evaluated. Section \ref{sec:result} provides a thorough discussion of the obtained results. Section \ref{sec:conclusion} concludes this paper by providing a guideline for possible future works. Finally, section \ref{sec:acknowledgment} appreciates everyone who supports and provides the chance to possible this research.

\section{Related Work}
\label{sec:related}
\subsection{Language Modelling}

Language modeling has gained popularity in recent years, and many works have been dedicated to building models for different languages based on varying contexts. Some works have sought to build character-level models. For example, a character-level model with Recurrent Neural Network (RNN) is presented in \cite{huang2019c}. This model reasons about word spelling and grammar dynamically. Another multi-task character-level attentional network model for the medical concept has been used to address Out-Of-Vocabulary (OOV) problem and to sustain morphological information inside the concept \cite{niu2019multi}.

Contextualized language modeling is centered around the idea that words can be represented differently based on the context in which they appear. Encoder-decoder language models, sequence autoencoders, and sequence-to-sequence models have this concept \cite{dai2015semi,ramachandran2016unsupervised,sutskever2014sequence}. ELMo and ULMFiT \cite{howard2018universal} are contextualized language models pre-trained on large general domain corpora. They are both based on LSTM networks \cite{hochreiter1997long}; ULMFiT benefits from a regular multi-layer LSTM network while ELMo utilizes a bidirectional LSTM structure to predict both next and previous words in a sequence of words. It then composes the final embedding for each token by concatenating the left-to-right and the right-to-left representations. Both ULMFiT and ELMo show considerable improvement in downstream tasks as compared to preceding language models and word embedding methods.

Another candidate for sequence-to-sequence mapping is the Transformer model \cite{vaswani2017attention}, which is based on the attention mechanism to evaluate dependencies between input/output sequences. Unlike LSTM, this model does not incorporate any recurrence. The Transformer model depends on two entities named encoder and decoder; the encoder takes the input sequence and maps it to a higher dimensional vector. This vector is then mapped to an output sequence by the decoder. Several pre-trained language modeling architectures are based on the transformer model, namely GPT \cite{radford2018improving} and BERT \cite{devlin2018bert}.

GPT includes a stack of twelve Transformer decoders. However, its structure is unidirectional, meaning that each token attends only to the previous one in the sequence. On the other hand, BERT performs joint conditioning on both left and right contexts by using a Masked Language Model (MLM) and a stack of transformer encoders along with the decoders. This way, BERT achieves an accurate pre-trained deep bidirectional representation.
There are other Transformer-based architectures such as XLNet \cite{yang2019xlnet}, RoBERTa \cite{liu2019roberta}, XLM \cite{lample2019cross}, T5 \cite{raffel2019exploring}, and ALBERT \cite{lan2019albert}, all of which have presented state-of-the-art results on multiple NLP tasks such as \cite{wang2018glue} and SQuAD \cite{rajpurkar2018know}.

Monolingual pre-trained models have been developed for several languages other than English. ELMo models are available for Portuguese, Japanese, German, and Basque \footnote{\url{https://allennlp.org/elmo}}. Regarding BERT-based models, BERTje for Dutch \cite{de2019bertje}, Alberto for Italian \cite{polignano2019alberto}, AraBERT for Arabic \cite{AraBert}, and other models for Finnish \cite{virtanen2019multilingual}, Russian \cite{kuratov2019adaptation} and Portuguese \cite{souza2019portuguese} have been released.

For the Persian language, several word embeddings such as Word2Vec, GloVe, and FastText \cite{grave2018learning} have been presented. All these word embeddings models are trained on Wikipedia corpus. A thorough comparison between these models is provided in \cite{zahedi2018persian} and shows that FastText and Word2Vec outperform other models. Another LSTM-based language model for Persian is presented in \cite{saravani2018persian}. Their model utilizes word embeddings as word representations and achieves the best performing model with a two-layer bidirectional LSTM network.

\subsection{NLP Downstream Tasks}
Although several works are presented to address NLP downstream tasks such as NER and Sentiment Analysis for the Persian language, the subject of pre-trained networks in the Persian language is a new topic. Most of the work done in this area is centered around machine learning or neural network methods built from scratch for each task, due to incapability of fine-tuning these approaches. For instance, a machine learning-based approach for Persian NER, using Hidden Markov Model (HMM), is presented in  \cite{ahmadi2015hybrid}. Another approach for Persian NER is provided by \cite{dashtipour2017persian} which combines a rule-based grammatical approach. Moreover, a Deep Learning approach for Persian NER is provided in \cite{bokaei2018improved} facilitating bidirectional LSTM networks. Beheshti-NER \cite{taher2020beheshti} uses multilingual Google BERT to form a fine-tuned model for Persian NER and is the closest work to present work. However, it only involves a fine-tuning phase for NER and does not entail developing a monolingual BERT-based model for the Persian Language.

The same situation applies to Persian sentiment analysis as Persian NER. In \cite{dastgheib2020application} a hybrid combination of Convolutional Neural Networks (CNN) and Structural Correspondence, Learning is presented to improve sentiment classification. Also, a graph-based text representation along with Deep Neural Learning is composed in \cite{bijari2020leveraging}. The closest work in sentiment analysis to the present work is DeepSentiPers \cite{sharami2020deepsentipers}, which leverages CNN and bidirectional LSTM networks combined with FastText trained over a balanced and augmented version of a Persian sentiment dataset known as SentiPers \cite{hosseini2018sentipers}.

It should be noted that none of these works uses pre-trained networks, and all of them focus solely on designing and combining methods to produce a task-specific approach.

\section{ParsBERT: Methodology}
\label{sec:methodology}

In this section, the methodology of our proposed model is presented. It consists of five main tasks, of which the first three concern the dataset and the next two concern model development. These tasks are data gathering, data pre-processing, accurate sentence segmentation, pre-training setup, and fine-tuning.

\subsection{Data Gathering}

Although a few Persian text corpora are provided by the University of Leipzig \cite{goldhahn-etal-2012-building} and University of Sorbonne \cite{ortizsuarez:hal-02148693}, the sentences in those corpora do not follow a logical corpora-level order and are somewhat erroneous. Also, these resources cover only a limited number of writing styles and subjects. Therefore, to increase the generality and efficiency of our pre-trained model in words, phrases, and sentence levels, it was necessary to compose a new form of the corpus from scratch to tackle the limitations mentioned earlier. This was done by crawling many sources such as Persian Wikipedia \footnote{\url{https://dumps.wikimedia.org/fawiki/}}, BigBangPage \footnote{\url{https://bigbangpage.com/}}, Chetor \footnote{\url{https://www.chetor.com/}}, Eligasht \footnote{\url{https://www.eligasht.com/Blog/}}, Digikala \footnote{\url{https://www.digikala.com/mag/}}, Ted Talks subtitles \footnote{\url{https://www.ted.com/talks}}, several fictional books and novels, and MirasText \cite{SABETI18.385}. The latter source has crawled more than 250 Persian news websites. Table \ref{table:1} demonstrates the statistics of our general-domain corpus:

\begin{table}[!htbp]
    \small	
    \centering
    \caption{Statistics and types of each source in the proposed corpus, entailing a varied range of written styles.}
    \label{table:1}
    % \begin{tabular*}{1.0\textwidth}{llll}
    \begin{tabularx}{1.0\linewidth}{p{0.005\textwidth}p{0.1\textwidth}p{0.18\textwidth}p{0.1\textwidth}}
        \toprule
        \# & Source & Type & Total Documents \\
        \midrule
        1 & Persian Wikipedia & General(encyclopedia) & 1,119,521 \\
        2 & BigBang Page & Scientific & 135 \\
        3 & Chetor & Lifestyle & 3,583 \\
        4 & Eligasht & Itinerary & 9,629 \\
        5 & Digikala & Digital magazine & 8,645 \\
        6 & Ted Talks & General (conversational) & 2,475 \\
        7 & Books & Novels, storybooks, short stories from old to the contemporary era & 13 \\
        8 & Miras-Text & News categories & 2,835,414 \\
        \hline
    \end{tabularx}
\end{table}

\subsection{Data Pre-Processing}
\begin{figure}[!ht]
\begin{subfigure}{0.45\textwidth}
\includegraphics[width=1.0\linewidth]{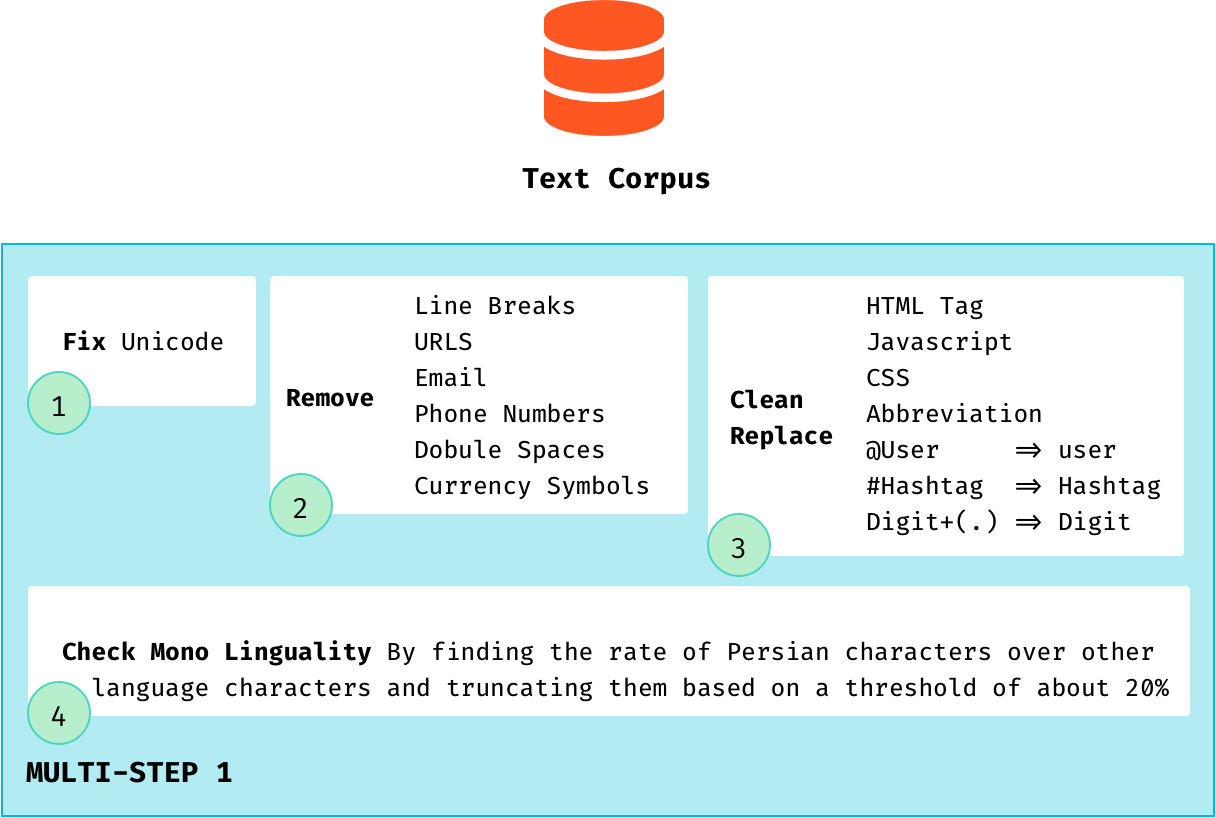} 
\caption{Step 1}
\label{fig:preprocessing_step_1_1}
\end{subfigure}\hspace{5mm}
\begin{subfigure}{0.45\textwidth}
\includegraphics[width=1.0\linewidth]{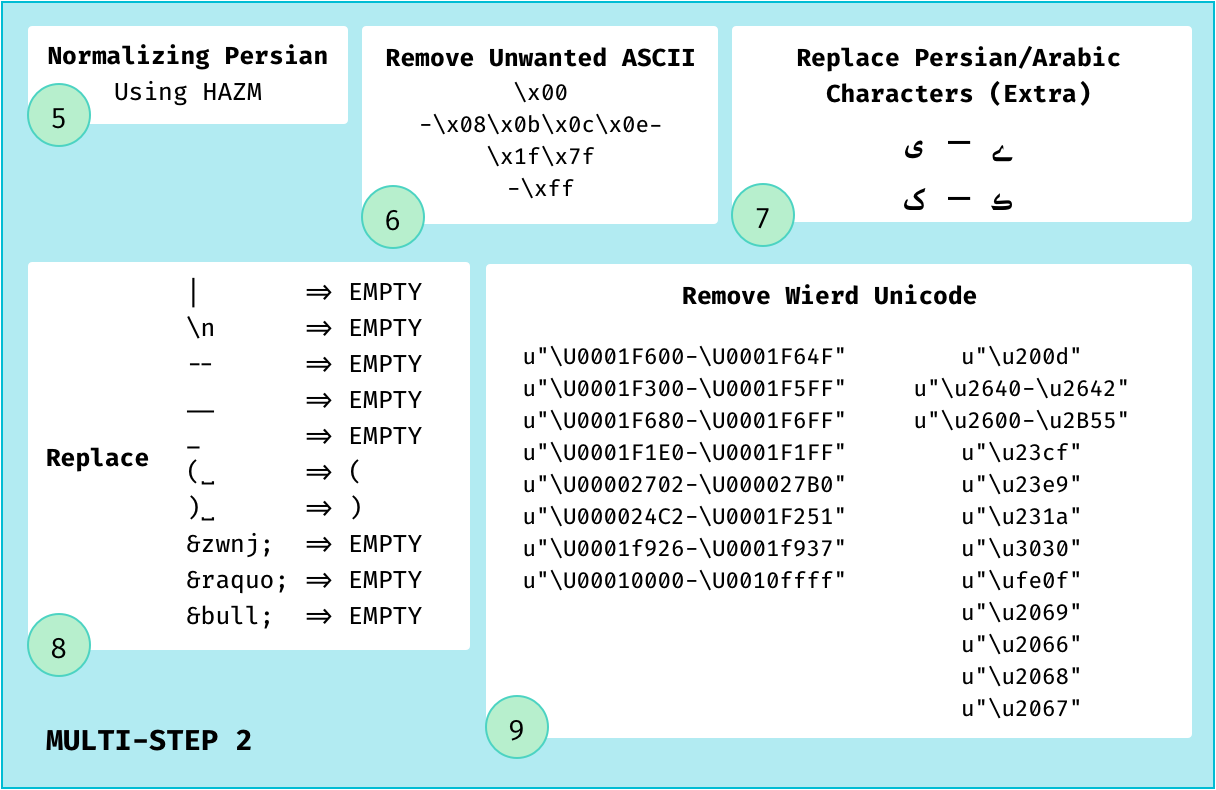}
\caption{Step 2}
\label{fig:preprocessing_step_1_2}
\end{subfigure}
\caption{Specific Persian corpus pre-processing that includes two steps: (a) removing all the trivial and junk characters and (b) standardizing the corpus with respect to Persian characters.}
\label{fig:preprocessing_steps}
\end{figure}

After gathering the pre-training corpus, an immense hierarchy of processing steps, including cleaning, replacing, sanitizing, and normalizing \footnote{\url{https://github.com/sobhe/hazm}}, is vital to transform the dataset into a proper format. This is done via a two-step process and is illustrated in Figure \ref{fig:preprocessing_steps}.

\subsection{Document Segmentation into True Sentences}
\begin{figure*}[]
\begin{subfigure}{1.0\textwidth}
\includegraphics[width=1.0\linewidth]{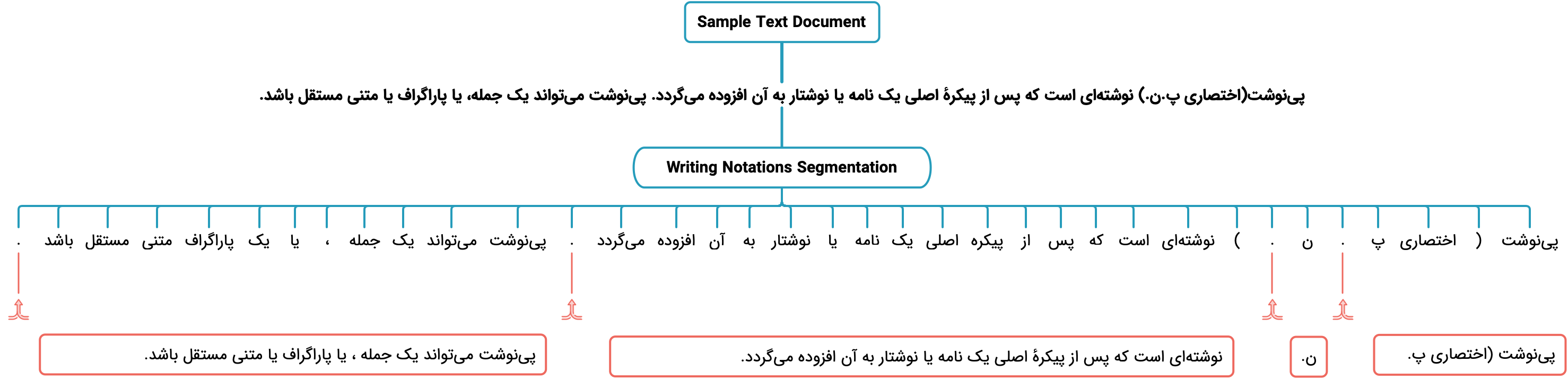} 
\caption{Notation Segmentation}
\label{fig:preprocessing_step_2_1}
\end{subfigure}\vspace{5mm}
\begin{subfigure}{1.0\textwidth}
\includegraphics[width=1.0\linewidth]{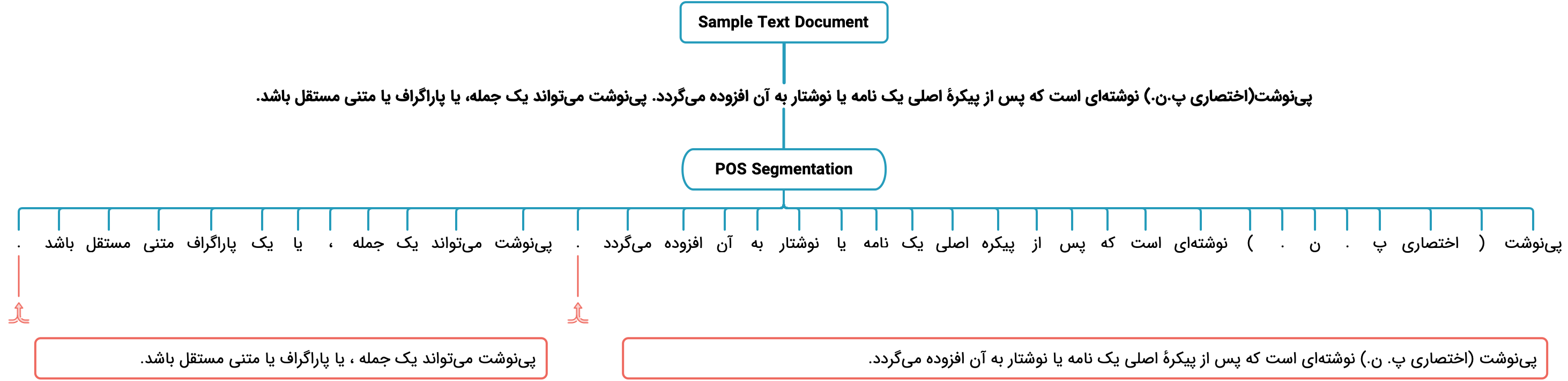}
\caption{POS Segmentation}
\label{fig:preprocessing_step_2_2}
\end{subfigure}
\caption{Example of segmenting a document into its sentences based on (a) only writing notations and (b) POS}
\label{fig:segmentation}
\end{figure*}

After the corpus is pre-processed, it should be segmented into True Sentences related to each document to achieve remarkable results for the pre-training model. A True Sentence in Persian is recognized based on this notations \textbf{[\RL{?!.:}]}. However, dividing content based merely on these notations has shown to cause problems. In Figure \ref{fig:segmentation}, an example of such issues is illustrated. It can be seen that the result includes short meaningless sentences without any vital information because there are abbreviations in Persian separated with the dot (.) notation. As an alternative, Part Of Speech (POS) can be a proper solution to handle these types of errors and to produce desired outputs.

This procedure enables the system to learn the real relationship between the sentences in each document. Table \ref{table:2} shows the statistics for the pre-training corpus segmented with the POS approach, resulting in 38,269,471 lines of True Sentences.

\begin{table}[!ht]
    \small	
    \centering
    \caption{Statistics of the pre-training corpus.}
    \label{table:2}
    \begin{tabularx}{1.0\linewidth}{llll}
        \toprule
        \# & Source & Total True Sentences \\
        \midrule
        1 & Persian Wikipedia & 1,878,008 \\
        2 & BigBang Page & 3,017 \\
        3 & Chetor & 166,312 \\
        4 & Eligasht & 214,328 \\
        5 & Digikala & 177,357 \\
        6 & Ted Talks & 46,833 \\
        7 & Books & 25,335 \\
        8 & Miras-Text & 35,758,281 \\
        \bottomrule
    \end{tabularx}
\end{table}

\subsection{Pre-training Setup}
Our model is based on BERT model architecture \cite{devlin2018bert}, which includes a multi-layer bidirectional Transformer. In particular, we use the original BERT \textsubscript{BASE} configuration: 12 hidden layers, 12 attention heads, 768 hidden sizes. The total number of parameters in this configuration is 110M. 
As per the original BERT pre-training objective, our pre-training objective consists of two tasks:

\begin{enumerate}
\item A Masked Language Model (MLM) is employed to train the model to predict randomly masked tokens by using cross-entropy loss. For this purpose given N tokens, 15\% of them are selected at random. From these selected tokens, 80\% of them are replaced by an exclusive [MASK] token, 10\% are replaced with a random token, and 10\% remain unchanged.

\item Implementing Next Sentence prediction (NSP) task, in which the model learns to predict whether the second sentence in a pair of sentences is the actual next sentence of the first one or not. In the original BERT paper \cite{devlin2018bert}, it has been argued that removing NSP from pre-training can attenuate the performance of the model on some tasks. Therefore, we employ NSP in our model to ensure high efficiency on different tasks.
\end{enumerate}

For model optimization \cite{kingma2014adam}, Adam optimizer with $\beta_1=0.9$ and $\beta_2=0.98$ is used for 1.9M steps. The batch size is set to 32, and each sequence contains 512 tokens at most. Finally, the learning rate is set to 1e-4.

Subword tokenization, which is necessary for better performance, is achieved using the WordPiece method \cite{kudo-2018-subword}. WordPiece operates as an intermediary between BPE \cite{sennrich-etal-2016-neural} and Unigram Language Model (ULM) approaches. WordPiece is trained on our pre-training corpus with a minimum frequency of three and 1.5K alphabet token limitations. The resulting vocabulary consists of 100K tokens, including unique BERT-specific tokens, namely [PAD], [UNK], [CLS], [MASK] [SEP] and $[\#\#]$ which is used as a prefix for word relation tokenization. Table \ref{table:3} shows an example of the tokenization process based on the WordPiece method. 

\begin{table*}
    \setfarsi
    \small	
    \centering
    \caption{Example of the segmentation process: (1) unsegmented sentence (2) segmented sentence using WordPiece method (\textvisiblespace \space interpret as -).}
    \label{table:3}
    \begin{tabularx}{1.0\linewidth}{m{0.9\linewidth}m{0.01\linewidth}}
        \toprule
        \\ 
        \begin{RLtext}
        brAy bAzdyd az dyw^c^smh bAyad bh nw^shr brawyd, ^shry kh az ^smAl bh dryAy xzr, az jnwb bh kwh\nospace hAy albrz, az ^sarq bh ^sahrstAn nwr w az .grb bh ^cAlws mnthy my\nospace ^swad.
        \end{RLtext}
        & (1) \\ \\ \hline
        \\
        
        \begin{RLtext}
        brAy  -- bAzdyd -- az -- dyw -- \#\#^c^smh -- bAyad -- bh -- nw^shr -- brawyd --, -- ^shry -- kh -- az -- ^smAl -- bh -- dryAy -- xzr -- , -- az -- jnwb -- bh -- kwh\nospace hAy -- albrz -- , -- az -- ^sarq -- bh -- ^sahrstAn -- nwr -- w -- az -- .grb -- bh -- ^cAlws -- mnthy -- my\nospace ^swad -- .
        \end{RLtext}
        & (2) \\ \\
        \bottomrule
    \end{tabularx}
\end{table*}

% \begin{table*}
%     \small	
%     \centering
%     \caption{Example of the segmentation process: (1) unsegmented sentence (2) segmented sentence using WordPiece method (\textvisiblespace \space interpret as **).}
%     \label{table:3}
%     \begin{tabularx}{1.0\linewidth}{m{0.9\linewidth}m{0.01\linewidth}}
%         \toprule
%         \\ \begin{Arabic}
%         \normalsize{
%         برای بازدید از دیوچشمه باید به نوشهر بروید، شهری که از شمال به دریای خزر، از جنوب به کوه‌های البرز، از شرق به شهرستان نور و از غرب به چالوس منتهی می‌شود.
%         }
%         \end{Arabic} & (1) \\ \\ \hline
%         \\ \begin{Arabic}
%         \normalsize{[CLS]  **  برای  **  بازدید  **  از  **  دیو  **  \#\#چشمه  **  باید  **  به  **  نوشهر  **  بروید  **  ،  **  شهری  **  که  **  از  **  شمال  **  به  **  دریای  **  خزر  **  ،  **  از  **  جنوب  **  به  **  کوههای  **  البرز  **  ،  **  از  **  شرق  **  به  **  شهرستان  **  نور  **  و  **  از  **  غرب  **  به  **  چالوس  **  منتهی  **  میشود  **  .  **  [SEP]}
%         \end{Arabic} & (2) \\ \\
%         \bottomrule
%     \end{tabularx}
% \end{table*}

\subsection{Fine-Tuning Setup}
The final language model (our proposed model) should be fine-tuned towards different tasks: Sentiment Analysis, Text Classification, and Named Entity Recognition. Sentiment Analysis and Text Classification belong to a broader task called Sequence Classification. Sentiment Analysis recognized as a specific task of Text Classification in representing the emotions behind the text.

\subsubsection{Sequence Classification}
Sequence classification is the process of labeling texts in a supervised manner. In our model, we incorporated the corresponding class for each sequence into the distinctive [CLS] token. We then added a simple feed-forward Softmax layer to predict the output classes. During this process, to maximize the log-probability of the correct class, both classifier and pre-trained model weights are adjusted.

\subsection{Named Entity Recognition}
This task aims to extract named entities in the text, such as names and label with appropriate NER classes such as locations, organizations, etc. The datasets used for this task contain sentences that are labeled with IOB format. In this format, tokens that are not part of an entity are tagged as "O", the "B" tag corresponds to the first word of an entity, and the "I" tag corresponds to the rest of the words of the same entity. Both "B" and "I" tags are followed by a hyphen (or underscore), followed by the entity category. Therefore, the NER task is a multi-class token classification problem that labels the tokens upon being fed a raw text. 

\section{Evaluation}
\label{sec:evaluation}
ParsBERT is evaluated on three downstream tasks: Sentiment Analysis (SA), Text Classification, and Named Entity Recognition (NER). Each of these tasks requires their specific datasets for the model to be fine-tuned and evaluated on.

\subsection{Sentiment Analysis}
It aims to classify text, such as comments based on their emotional bias. The proposed model is evaluated on three sentiment datasets as follows:

\begin{enumerate}
\item Digikala user comments provided by Open Data Mining Program \footnote{\url{https://www.digikala.com/opendata/}} (ODMP). This dataset contains 62,321 user comments with three labels: (0) No Idea, (1) Not Recommended and (2) Recommended.
\item Snappfood \footnote{\url{https://snappfood.ir/}} (an online food delivery company) user comments containing 70,000 comments with two labels (i.e. polarity classification): (0) Happy and (1) Sad.
\item DeepSentiPers \cite{sharami2020deepsentipers}, which is a balanced and augmented version of SentiPers \cite{hosseini2018sentipers}, contains 12,138 user opinions about digital products labeled with five different classes; two positives (i.e., happy and delighted), two negatives (i.e., furious and angry) and one neutral class. Therefore, this dataset can be utilized for both multi-class and binary classification. In the case of binary classification, the neutral class and its corresponding sentences are removed from the dataset.
\end{enumerate}

\begin{figure*}
\begin{subfigure}{0.45\textwidth}
\includegraphics[width=1.0\linewidth]{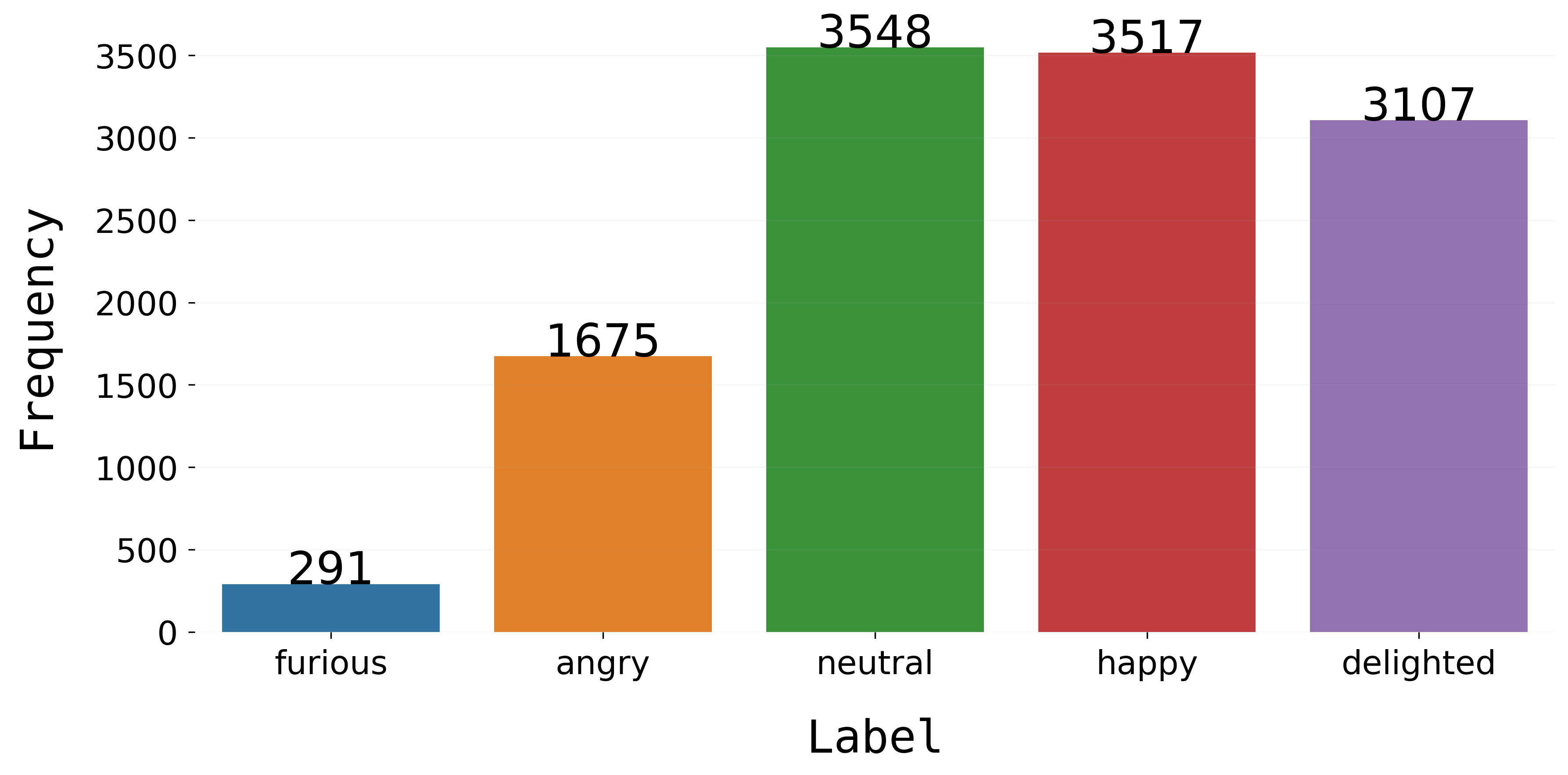} 
\caption{}
\label{fig:deepsentipersemulti}
\end{subfigure}\hspace{10mm}
\begin{subfigure}{0.45\textwidth}
\includegraphics[width=1.0\linewidth]{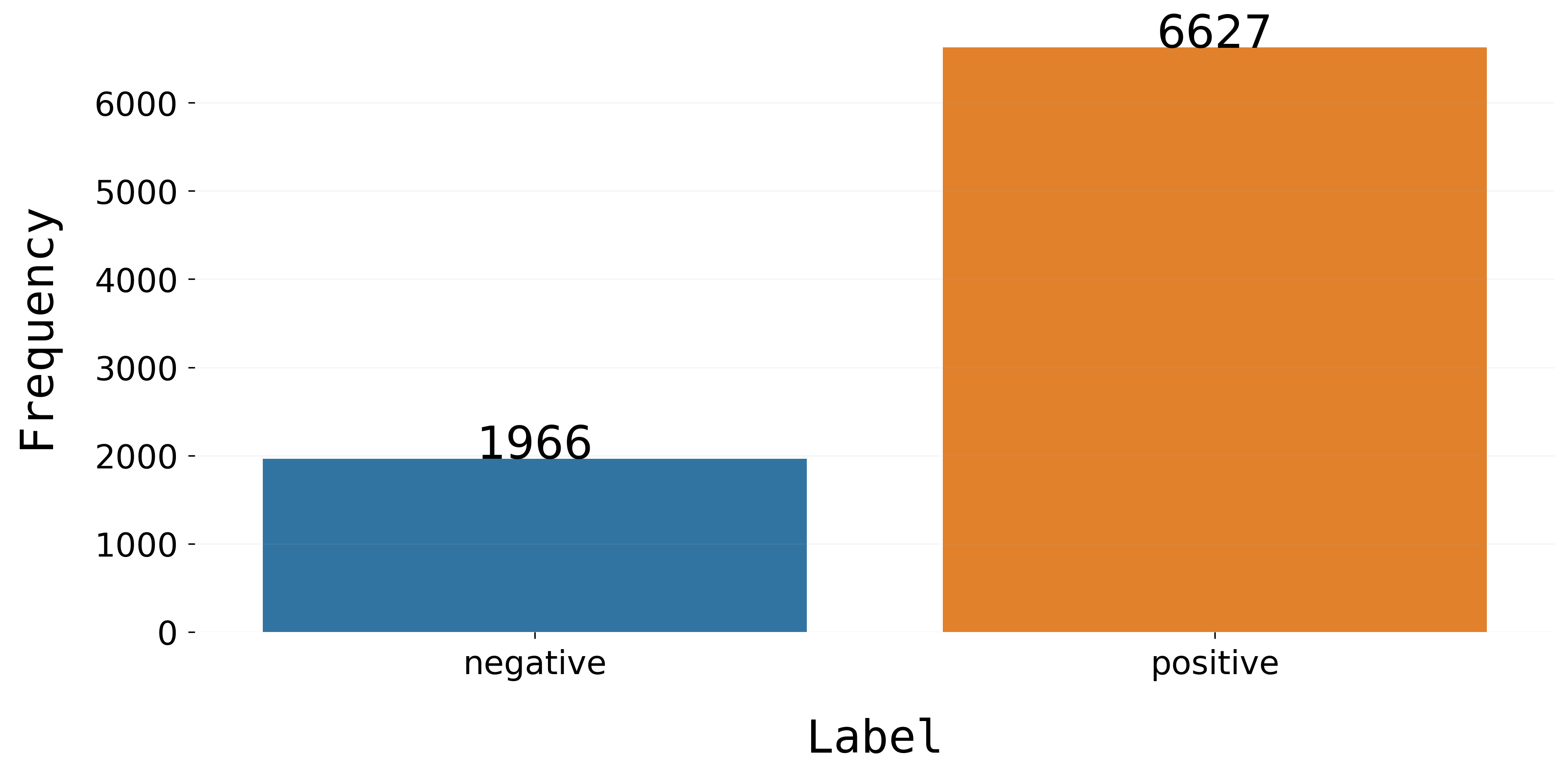}
\caption{}
\label{fig:deepsentipersbinary}
\end{subfigure}
\begin{subfigure}{0.45\textwidth}
\includegraphics[width=1.0\linewidth]{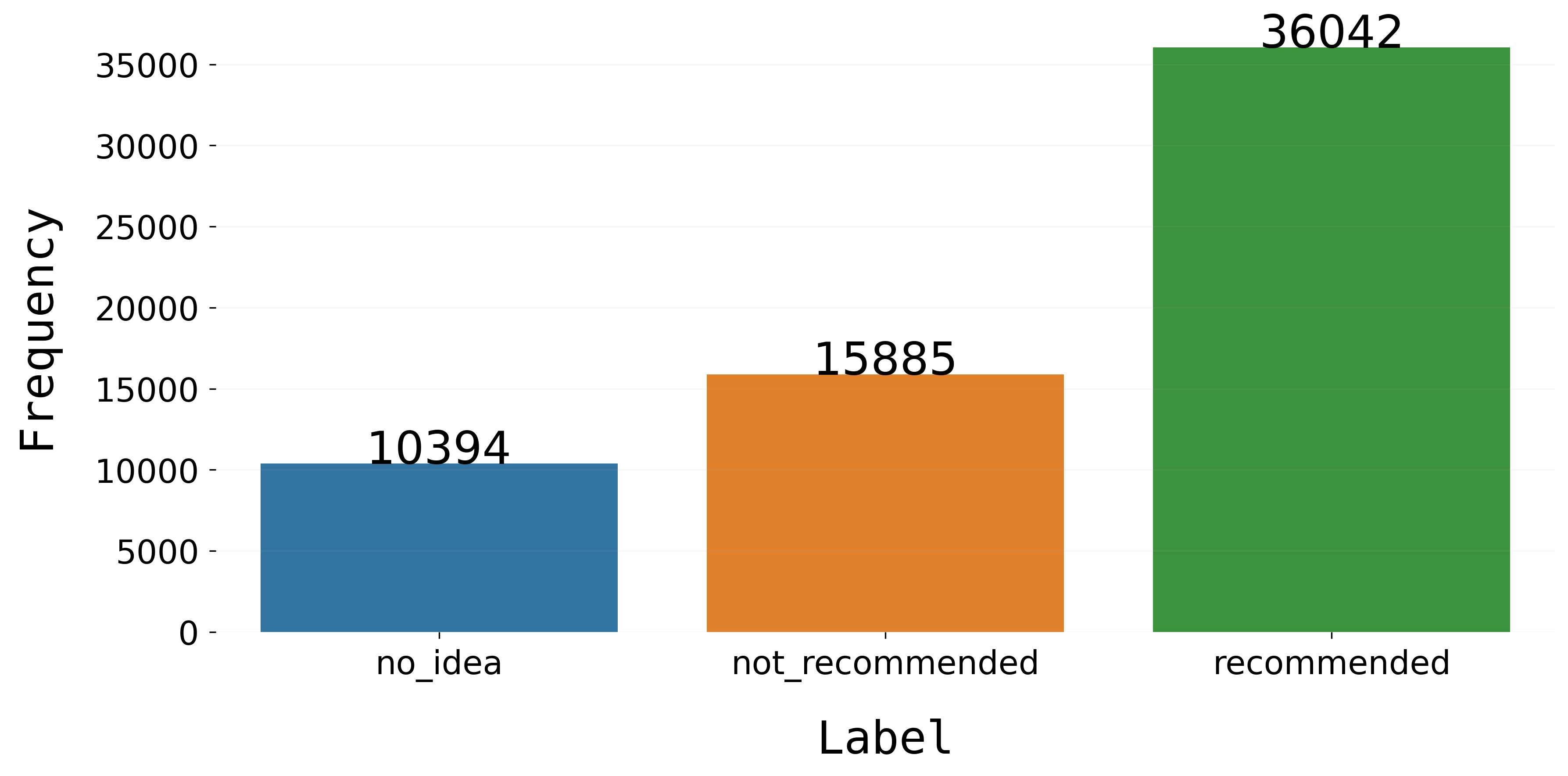}
\caption{}
\label{fig:sentimentdigikala}
\end{subfigure}\hspace{17mm}
\begin{subfigure}{0.45\textwidth}
\includegraphics[width=1.0\linewidth]{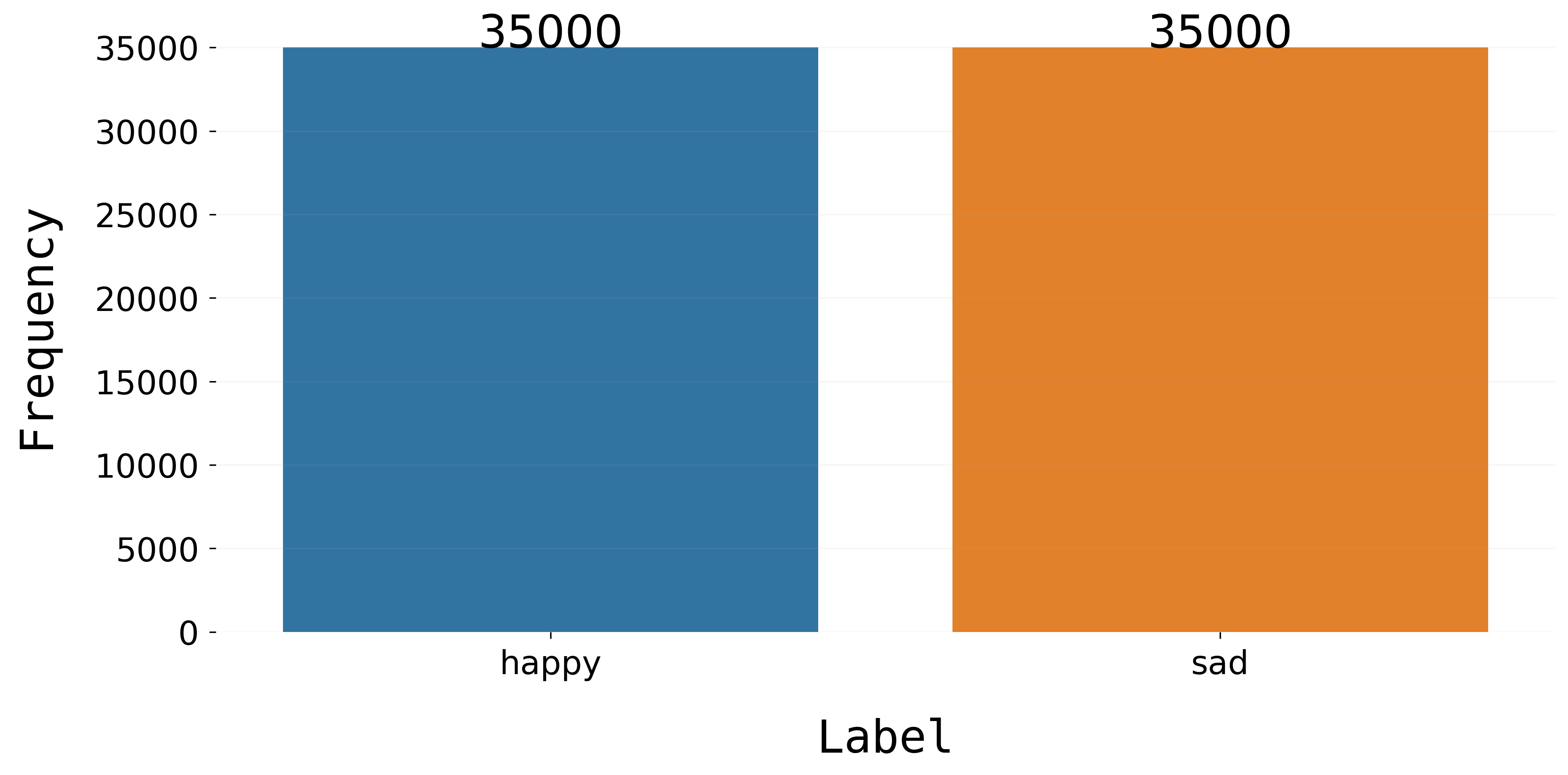}
\caption{}
\label{fig:sentimentsnapp}
\end{subfigure}
\caption{Class distribution for (a) Multi-class DeepSentiPers, (b) Binary-class DeepSentiPers, (c) Digikala and (d) SnappFood datasets.}
\label{fig:sentimentchart}
\end{figure*}

The second dataset of the above list was not readily available. We extracted it using our tools to provide a more comprehensive evaluation. Figure \ref{fig:sentimentchart} illustrates the class distribution for all three sentiment datasets.

\textbf{Baselines:} Since no work has been done regarding the Digikala and SnappFood datasets, our baseline for these datasets is the multilingual BERT model. As for the DeepSentiPers \cite{sharami2020deepsentipers} dataset, we compare our results with those reported in this paper. Their methodology for addressing the SA task entails a hybrid CNN and BiLSTM networks.

\subsection{Text Classification}
Text classification is an important NLP task in which the objective is to classify a text-based on pre-determined classes. The number of classes is usually higher than that of sentiment analysis and words distribution makes finding the right and main class so tricky. The datasets used for this task come from two sources:

\begin{enumerate}
\item A total of 8,515 articles scraped from Digikala online magazine \footnote{\url{https://www.digikala.com/mag/}}. This dataset includes seven different classes. 
\item A dataset of various news articles scraped from different online news agencies' websites. The total number of articles is 16,438, spread over eight different classes.
\end{enumerate}

\begin{figure*}
\begin{subfigure}{0.45\textwidth}
\includegraphics[width=1.0\linewidth]{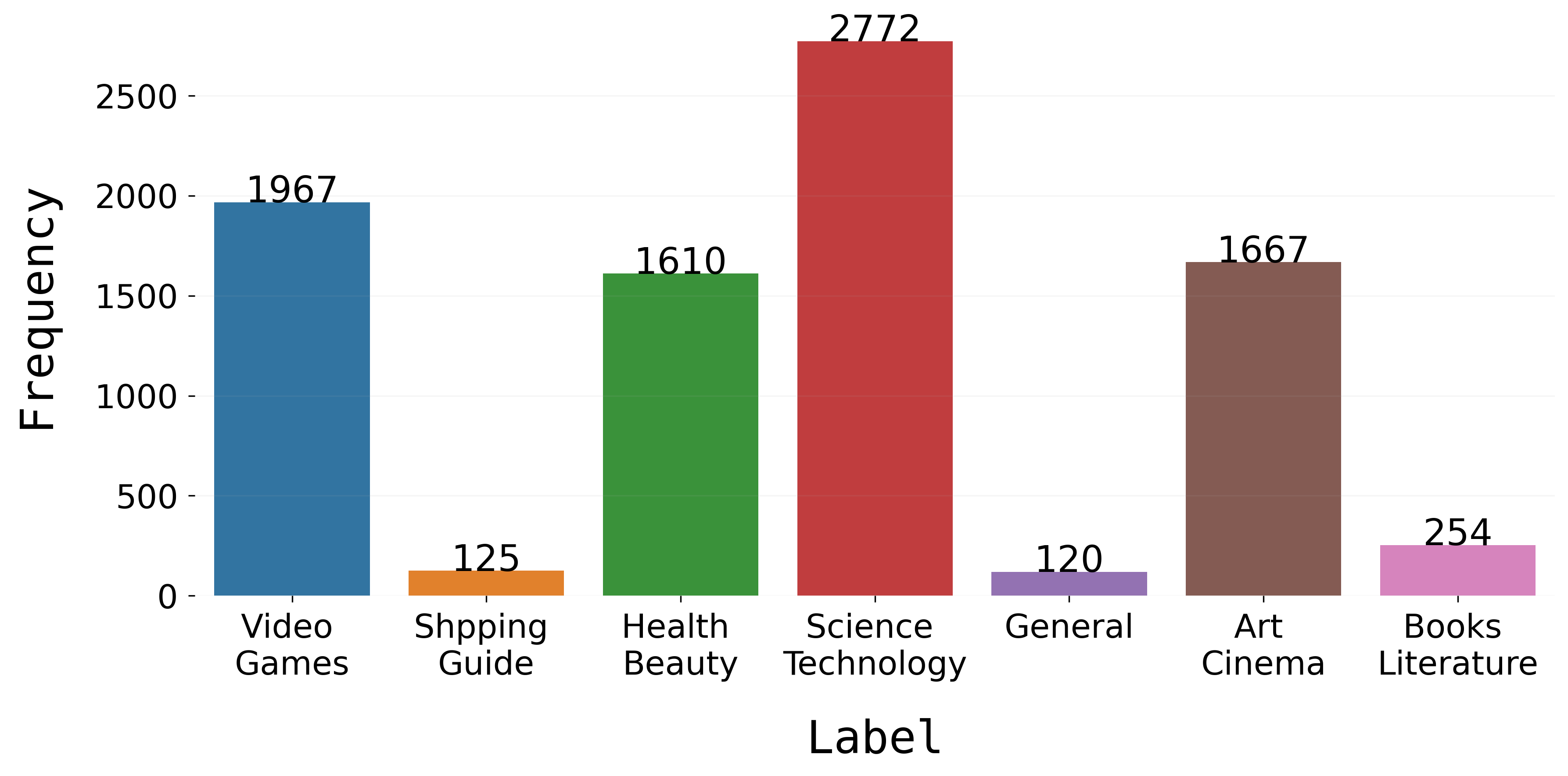} 
\caption{}
\label{fig:digikalamag}
\end{subfigure}\hspace{10mm}
\begin{subfigure}{0.45\textwidth}
\includegraphics[width=1.0\linewidth]{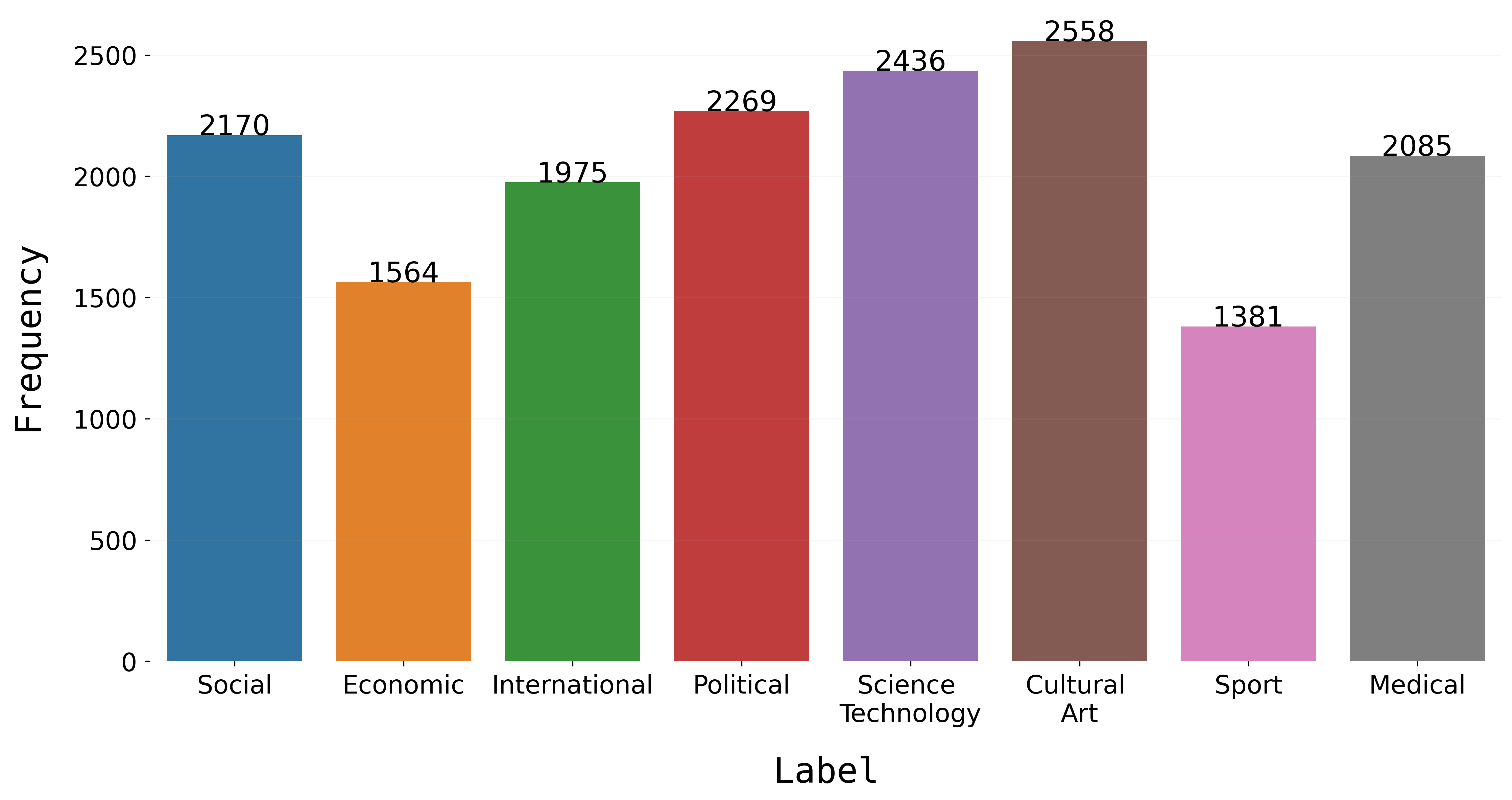}
\caption{}
\label{fig:persiannews}
\end{subfigure}
\caption{Class distribution for (a) Digikala Online Magazine and (b) Persian news articles scraped from various websites.}
\label{fig:textclf}
\end{figure*}

We have scraped and prepared both of these datasets using our own tools. Figure \ref{fig:textclf} shows the class distribution for each of these datasets.

\textbf{Baseline:} Since we have prepared both datasets for this task using our tool, no prior work has been done. Therefore, we only have the monolingual BERT model to compare our model to for this task.

\subsection{Named Entity Recognition}
\begin{figure}[!htbp]
\begin{subfigure}{0.45\textwidth}
\vspace{4mm}
\includegraphics[width=1.0\linewidth]{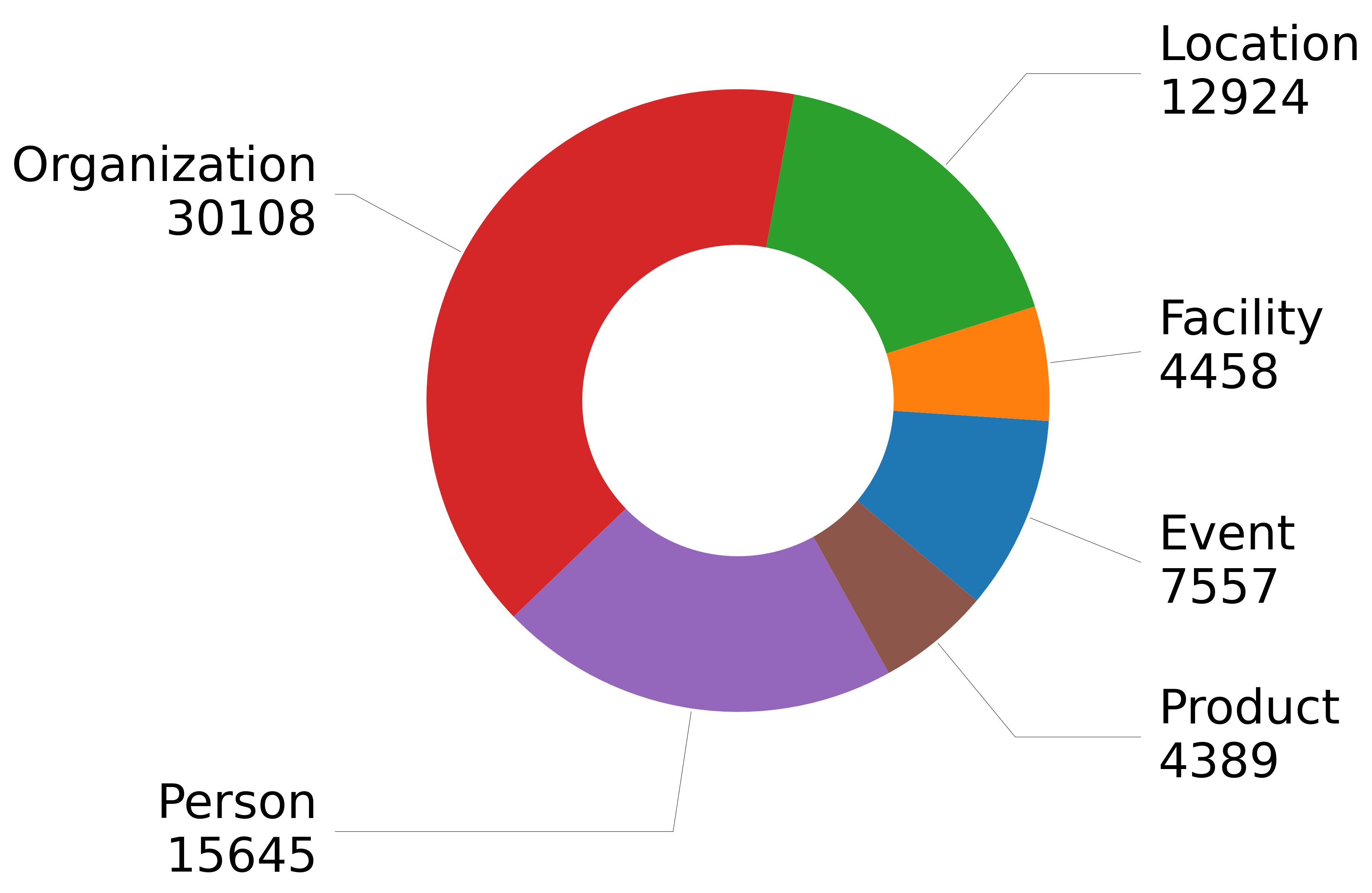} 
\caption{}
\label{fig:nerarman}
\end{subfigure}\hspace{5mm}
\begin{subfigure}{0.45\textwidth}
\includegraphics[width=1.0\linewidth]{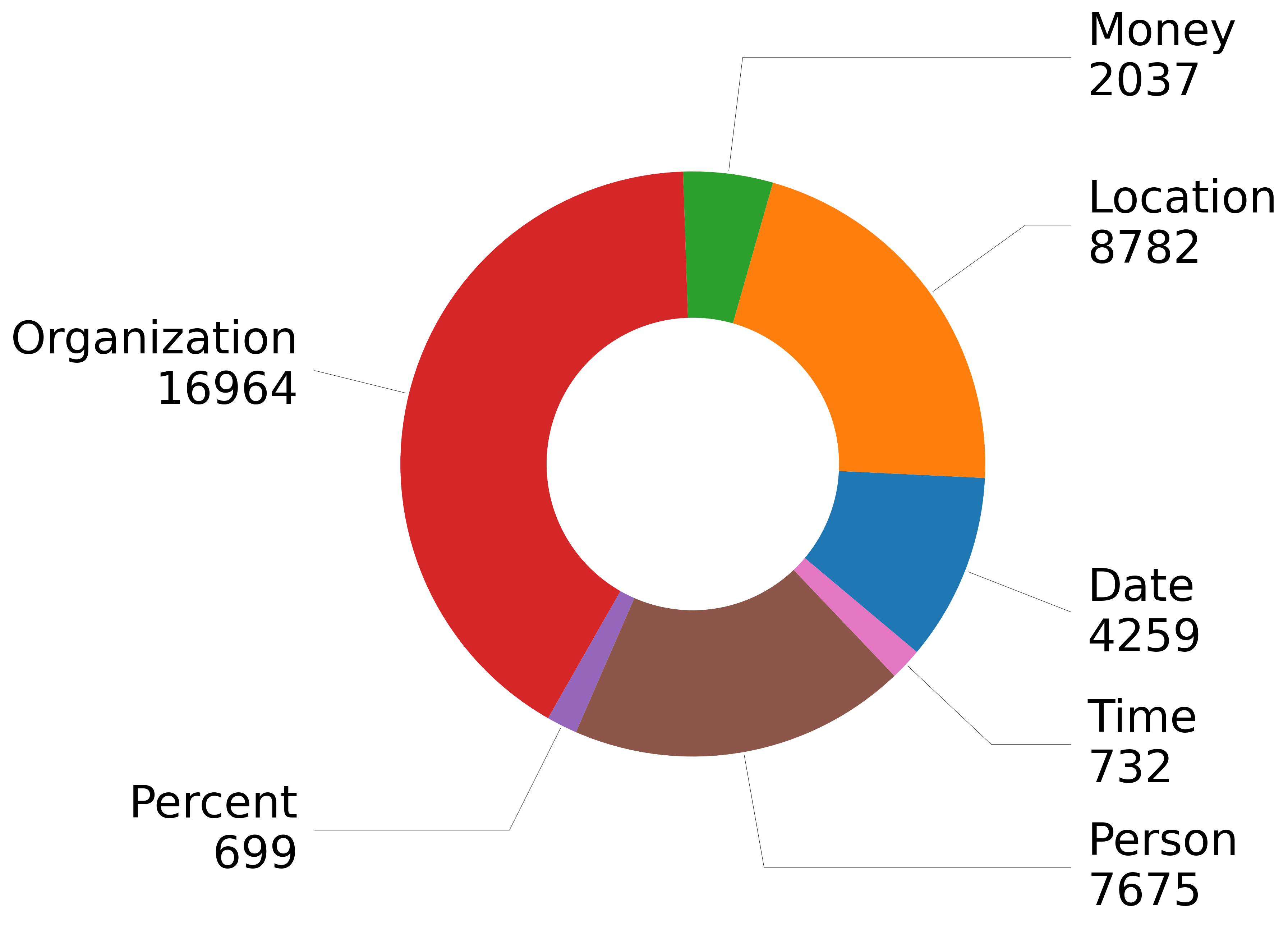}
\caption{}
\label{fig:nerpeyma}
\end{subfigure}
\caption{Class distribution for (a) ARMAN and (b) PEYMA datasets}
\label{fig:nerchart}
\end{figure}

For the NER task evaluation, PEYMA \cite{shahshahani2018peyma} and ARMAN \cite{poostchi2018bilstm} readily available datasets are used. PEYMA dataset includes 7,145 sentences with a total of 302,530 tokens from which 41,148 tokens are tagged with seven different classes. On the other hand, the ARMAN dataset holds 7,682 sentences with 250,015 sentences tagged over six different classes. The class distribution for these datasets is shown in Figure \ref{fig:nerchart}.

\textbf{Baselines:} We compare the result of our model for the NER task to that of Beheshti-NER \cite{taher2020beheshti}. Beheshti-NER utilizes a multilingual BERT model to tackle the same NER task as ours. 

\section{Results}
\label{sec:result}

\subsection{Sentiment Analysis Results} 
Table \ref{table:sentiment_digikala_snappfood_results} shows the results obtained on Digikala and SnaooFood datasets. This table shows that ParsBERT outperforms the multilingual BERT model in terms of accuracy and $F_{1}$ score.

\begin{table}[!htbp]
    \small	
    \centering
    \caption{ParsBERT performance on Digikala and SnappFood datasets compared to multilingual BERT model.}
    \label{table:sentiment_digikala_snappfood_results}
    % \begin{tabular*}{1.0\textwidth}{lllll}
    \begin{tabularx}{1.0\linewidth}{lcccc}
    % \begin{tabular}{{|c|cc|cc|}}
        \toprule
         \multirow{2}{*}{Model} & \multicolumn{2}{c}{Digikala} & \multicolumn{2}{c}{SnappFood} \\
         & $Acuuracy$ & $F_{1}$ & $Acuuracy$ & $F_{1}$ \\
         \midrule
         ParsBERT & \textbf{82.52} & \textbf{81.74} & \textbf{87.80} & \textbf{88.12} \\ 
         multilingualBERT & 81.83 & 80.74 & 87.44 & 87.87
    \end{tabularx}
\end{table}

The results for DeepSentiPers dataset are presented in table \ref{table:sentiment_sentipers_results}. It can be seen that ParsBERT achieves significantly higher $F_{1}$ scores for both multi-class and binary sentiment analysis compared to methods mentioned in DeepSentiPers \cite{sharami2020deepsentipers}.

\begin{table}[!htbp]
    \small	
    \centering
    \caption{ParsBERT performance on DeepSentiPers dataset compared to methods mentioned in DeepSentiPers \cite{sharami2020deepsentipers}}
    \label{table:sentiment_sentipers_results}
    % \begin{tabular*}{1.0\textwidth}{lllll}
    \begin{tabularx}{1.0\linewidth}{lcc}
    % \begin{tabular}{{|c|cc|cc|}}
        \toprule
         \multirow{2}{*}{Model} & \multicolumn{1}{c}{Multi-Class} & \multicolumn{1}{c}{Binary} \\
         & $F_{1}$ & $F_{1}$ \\
         \midrule
         ParsBERT & \textbf{71.11} & \textbf{92.13} \\ 
         CNN + FastText \cite{sharami2020deepsentipers} & 66.30 & 80.06 \\ 
         CNN \cite{sharami2020deepsentipers} & 66.65 & 91.90 \\ 
         BiLSTM + FastText \cite{sharami2020deepsentipers} & 69.33 & 90.59 \\ 
         BiLSTM \cite{sharami2020deepsentipers} & 66.50 & 91.98 \\ 
         SVM \cite{sharami2020deepsentipers} & 67.62 & 91.31
    \end{tabularx}
\end{table}

\subsection{Text Classification Results} 
The obtained results for text classification task are summarized in Table \ref{table:text_clf_results}. It can be seen that ParsBERT achieves better accuracy and scores compared to multilingual BERT model on both Digikala Magazine and Persian news datasets.

\begin{table}[!htbp]
    \small	
    \centering
    \caption{ParsBERT performance on text classification task compared to multilingual BERT model.}
    \label{table:text_clf_results}
    % \begin{tabular*}{1.0\textwidth}{lllll}
    \begin{tabularx}{1.0\linewidth}{lcccc}
    % \begin{tabular}{{|c|cc|cc|}}
        \toprule
         \multirow{2}{*}{Model} & \multicolumn{2}{c}{Digikala Magazine} & \multicolumn{2}{c}{Persian News} \\
         & $Acuuracy$ & $F_{1}$ & $Acuuracy$ & $F_{1}$ \\
         \midrule
         ParsBERT & \textbf{94.28} & \textbf{93.59} & \textbf{97.20} & \textbf{97.19} \\ 
         multilingualBERT & 91.31 & 90.72 & 95.80 & 95.79
    \end{tabularx}
\end{table}

\subsection{Named Entity Recognition Results} 
Obtained results for NER task indicates that ParsBERT outperforms all prior works in this area by achieving $F_{1}$ scores as high as 93.10 and 98.79 for PEYMA and ARMAN datasets, respectively. A thorough comparison between ParsBERT performance and other works on these two datasets is provided in table \ref{table:ner_results}. 

\begin{table}[!htbp]
    \small	
    \centering
    \caption{ParsBERT performance on PEYMA and ARMAN datasets for the NER task compared to prior works.}
    \label{table:ner_results}
    % \begin{tabular*}{1.0\textwidth}{lllll}
    \begin{tabularx}{1.0\linewidth}{lcc}
    % \begin{tabular}{{|c|cc|cc|}}
        \toprule
         \multirow{2}{*}{Model} & \multicolumn{1}{c}{PEYMA} & \multicolumn{1}{c}{ARMAN} \\
         & $F_{1}$ & $F_{1}$ \\
         \midrule
         ParsBERT & \textbf{93.10} & \textbf{98.79} \\ 
         MorphoBERT \cite{Taghizadeh2020NSURL2019T7} & - & 89.9 \\ 
         Beheshti-NER \cite{taher2020beheshti} & 90.59 & 84.03 \\ 
         LSTM-CRF \cite{Hafezi2018} & - & 86.55 \\ 
         Rule-Based-CRF \cite{shahshahani2018peyma} & 84.00 & - \\ 
         BiLSTM-CRF \cite{poostchi2018bilstm} & - & 77.45 \\ 
         LSTM \cite{Hafezi2018} & - & 73.61 \\ 
         Deep CRF \cite{bokaei2018improved} & - & 81.50 \\
         Deep Local \cite{bokaei2018improved} & - & 79.10 \\
         SVM-HMM \cite{poostchietal2016personer} & - & 72.59
    \end{tabularx}
\end{table}

\subsection{Discussion}
ParsBERT successfully achieves state-of-the-art performance on all mentioned downstream tasks. This conclusively proves that monolingual language models outmatch multilingual ones. In the case of ParsBERT, this improvement roots in several causes. Firstly, the standardization and pre-processing employed in the current methodology overcomes the lack of correct sentences in Persian corpora and takes into account the complexities of the Persian language. Secondly, the range of topics and writing styles included in the pre-training dataset is much more diverse than that of multilingual BERT that only applies the Wikipedia dataset. Another limitation of the multilingual model caused by using the small Wikipedia corpus is that it contains a vocab size of 70K tokens for all 100 languages it supports. ParsBERT, on the other hand, incorporates a 14GB corpus with more than 3.9M documents with a vocab size of 100K. All in all, the obtained results indicate that ParsBERT is more competent at perceiving and understanding the Persian language than multilingual BERT or any of the previous works that have followed the same objective.

\section{Conclusion}
\label{sec:conclusion}
There are few specific language models for the Persian language capable of providing state-of-the-art performance on different NLP tasks. ParsBERT is a fresh model that is lighter than multilingual BERT and represents state-of-the-art results in downstream tasks, such as Sentiment Analysis, Text Classification, and Named Entity Recognition. 
Compared to other Persian NER competitor models, ParsBERT outperforms all prior works in terms of $F_{1}$ score by achieving 93\% and 98\% scores for PEYMA and ARMAN datasets, respectively. Moreover, in the SA task, ParsBERT gained better performance on the SentiPers dataset against the DeepSentiPers model by achieving $F_{1}$ scores as high as 92\% and 71\% for both binary and multi-label scenarios. In all cases, ParsBERT outperforms multilingual BERT and other suggestion networks.

The number of datasets for downstream tasks in Persian is limited. Therefore, we composed a considerable set of datasets to evaluate ParsBERT performance on them. These datasets will soon be published for public use \footnote{\url{https://hooshvare.github.io/}}. Also, we happily announce that ParsBERT synchronizes to Huggingface Transformers \footnote{\url{https://huggingface.co/}} for any public use and to serve as a new baseline for numerous Persian NLP use cases \footnote{\url{https://github.com/hooshvare/parsbert}}.

\section{Acknowledgments}
\label{sec:acknowledgment}
We hereby, express our gratitude to the Tensorflow Research Cloud (TFRC) program\footnote{\url{https://tensorflow.org/tfrc}} for providing us with the necessary computation resources. We also thank Hooshvare\footnote{\url{https://hooshvare.com}} Research Group for facilitating dataset gathering and scraping online text resources.

\normalsize
\bibliography{references}

\end{document}